\documentclass{article}
\usepackage{spconf,amsmath,epsfig}
\usepackage{amsfonts}%Added by us
\usepackage{fixltx2e}%Added by us
\usepackage{color}% our
\usepackage{framed,multirow}% our
\usepackage{algpseudocode}%Added by us
\usepackage{algorithm}%Added by us
\usepackage[caption=false,font=normalsize,labelfont=sf,textfont=sf]{subfig}
\title{Hyperspectral Image Restoration via Global Total Variation Regularized Local nonconvex Low-Rank matrix Approximation}
\name{ Haijin Zeng, Xiaozhen Xie*\thanks{*Corresponding author: xiexzh@nwafu.edu.cn. }, Jifeng Ning\thanks{This work was supported by the Fundamental Research Funds for the Central Universities under Grant No.
  		2452019073 and the National Natural Science Foundation of
  		China under Grant No. 61876153.}}
\address{ Northwest A\&F University, College of Science, Yangling 712100, P.R. China
}
\begin{document}
%\ninept
%

\maketitle
\begin{abstract}
Several bandwise total variation (TV) regularized low-rank (LR)-based models have been proposed to remove mixed noise in hyperspectral images (HSIs).
Conventionally, the rank of LR matrix is approximated using nuclear norm (NN).
The NN is defined by adding all singular values together, which is essentially a $L_1$-norm of the singular values.
It results in non-negligible approximation errors and thus the resulting matrix estimator can be significantly biased.
Moreover, these bandwise TV-based methods exploit the spatial information in a separate manner.
To cope with these problems, we propose a spatial–spectral TV (SSTV) regularized non-convex local LR matrix approximation (NonLLRTV) method to remove mixed noise in HSIs.
From one aspect, local LR of HSIs is formulated using a non-convex $L_{\gamma}$-norm, which provides a closer approximation to the matrix rank than the traditional NN.
From another aspect, HSIs are assumed to be piecewisely smooth in the global spatial domain.
The TV regularization is effective in preserving the smoothness and removing Gaussian noise.
These facts inspire the integration of the NonLLR with TV regularization.
To address the limitations of bandwise TV, we use the SSTV regularization to simultaneously consider global spatial structure and spectral correlation of neighboring bands.
Experiment results indicate that the use of local non-convex penalty and global SSTV can boost the preserving of spatial piecewise smoothness and overall structural information.
\end{abstract}
\section{Introduction}
\label{sec:intro}

Hyperspectral images (HSIs) can provide spectral information of hundreds of continuous bands in the same scene.
They are widely used in many fields. In recent years, HSIs have attracted great research interest in the field of remote sensing.
However, due to the limitations of observation conditions and sensors, the HSI obtained by hyperspectral imagers is usually contaminated by a variety of noises, such as Gaussian noise, stripes, deadlines, and impulse noise. These noises adversely affect the image quality of HSIs, the subsequent processing and applications.
%Therefore, as a pre-processing step, HSI restoration is an important research direction that needs to be further studied.

Low rank (LR) model is a powerful tool in image processing, and its purpose is to decompose the observation data into a low rank matrix representing ideal data and a sparse matrix representing sparse noise.
Based on LR model, numerous approaches have been proposed for HSI restoration.
Albeit the success of LR models in theoretical research and practical applications,
they may obtain suboptimal performance in real applications,
since the nuclear norm (NN) may not be a good approximation to the rank function.
Specifically, compared to the rank function in which all the nonzero singular values have equal contributions,
the NN treats the singular values differently by adding them together.
Moreover, the theoretical requirements (e.g., incoherence property) of the NN are usually very hard to satisfy in practical scenarios.
Recently, a number of studies, both practically and theoretically, have shown that non-convex penalty of LR can provide better estimation accuracy
and variable selection consistency than NN \cite{nonconvex_optimal}.
Motivated by such facts, several non-convex penalties have been proposed and studied as alternatives to NN.

In this paper, we propose a global spatial-spectral total variation (SSTV) regularized local non-convex LR matrix approximation (NonLLRTV) method for HSI denoising.
Specifically, the HSIs are first divided into overlapping patches.
Then, from one aspect, the clean HSI patches have its underlying local LR property,
even though the observed HSI data may not be globally LR due to outliers and non-Gaussian noise.
According to this fact, in our model, the local LR of hyperspectral data is represented by the newly emerged nonconvex $L_{\gamma}$-norm \cite{Non_LRMA},
which provides a closer approximation to the matrix rank than the traditional NN.
From another aspect, HSIs are assumed to be piecewisely smooth in the global spatial domain.
The TV regularization is effective in preserving the spatial piecewise smoothness and removing Gaussian noise.
These facts inspire the integration of the NonLLR with TV regularization.
To address the limitations of bandwise TV, we use the SSTV regularization to simultaneously consider global spatial structure and spectral correlation of neighboring bands.

\section{Problem formulation} \label{key model}
%\subsection{Local Low-Rank matrix  approximation-based HSI Restoration}
On the context of HSIs, it is well known that each spectral characteristic can be represented by a linear combination of a small number of pure spectral endmembers.
It means that the Casorati matrix (a matrix whose columns comprise vectorized bands of the HSI) $\mathbf{L}$ of clean HSI $\mathcal{L}$ can be decomposed into $\mathbf{L} = \mathbf{U}\mathbf{V}$.
Then, the image degradation model can be expressed as $\mathbf{O=UV+S+N}$,
where $\mathbf{O}$, $\mathbf{S}$, $\mathbf{N}$ denote the observed HSI, sparse noise and Gaussian noise, respectively.

Unfortunately, matrix $\mathbf{O}$ is a morbid matrix with a huge difference in the number of columns and rows, i.e., $mn \gg p$, which result in blurring and a loss of details. To alleviate this problem and effectively explore the local low rank structure of underlying HSI, we denoise HSI patch by patch and first define an operator $\mathrm{P}_{i, j}:\mathcal{L} \rightarrow \mathbf{L}_{i, j}$.
This binary operator $\mathrm{P}_{i, j}$ is used to extract $m_1 \times n_1 $ rows from HSI data $\mathcal{L} \in \mathbb{R}^{m \times n \times p}$, i.e., $\mathcal{L}_{i, j}=\mathrm{P}_{i, j}(\mathcal{L})=\mathbf{L}_{i, j}$,
where the spatial size of $m_1 \times n_1$ is centralized at pixel $(i, j)$ of HSI data, $m_1n_1$ is approximately equal to $p$.
$\mathrm{P}_{i, j}^{\text{T}}$ is the inverse of $\mathrm{P}_{i, j}$.

%
%In this section, we first define an operator $\mathrm{P}_{i, j}:\mathcal{L} \rightarrow \mathcal{L}_{i, j}$.
%This binary operator $\mathrm{P}_{i, j}$ is used to extract a 3-D patch $\mathcal{L}_{i, j} \in \mathbb{R}^{m_1 \times n_1 \times p}$ from HSI data $\mathcal{L} \in \mathbb{R}^{m \times n \times p}$,
%where the spatial size of $m_1 \times n_1$ is centralized at pixel $(i, j)$ of HSI data.
%$\mathrm{P}_{i, j}^{\text{T}}$ is the inverse of $\mathrm{P}_{i, j}$.
%By the above definition, we apply the operation $\mathrm{P}_{i, j}$ on the tensors $\mathcal{O}$, $\mathcal{L}$, $\mathcal{S}$, $\mathcal{N}$. %divide them into overlapping patches as following
%Then, for a patch centered at the pixel $(i, j)$, the image degradation model is expressed as $\mathbf{O}_{i,j}=\mathbf{U}_{i,j}\mathbf{V}_{i,j}+\mathbf{S}_{i,j}+\mathbf{N}_{i,j}$.

As \cite{LRTV}, we assume that each element of $\mathcal{U}_{i,j} \in \mathbb{R}^{m_{1}n_{1} \times r},\\ \mathcal{V}_{i,j} \in \mathbb{R}^{r \times p} $ is sampled from the Gaussian distribution, the sparse error $\mathcal{S}_{i,j}$ is sampled from the Laplace distribution, and the noise $\mathcal{G}_{i,j}$ obeys a Gaussian distribution, i.e., $\mathcal{U}_{i,j} \sim \mathcal{N}\left(0, \lambda_{u}^{-1}\right)$, $\mathcal{V}_{i,j} \sim \mathcal{N}\left(0, \lambda_{v}^{-1}\right)$, $\mathcal{S}_{i,j} \sim\mathcal{L}\left(0, \lambda_{s}^{-1}\right)$, $\mathcal{N}_{i,j} \sim \mathcal{N}\left(0, \lambda_{g}^{-1}\right)$.
By treating $\mathcal{U}_{i,j}, \mathcal{V}_{i,j},$ and $\mathcal{S}_{i,j}$ as model parameters, and $\lambda_{u}, \lambda_{v}$ $\lambda_{s},$ and $\lambda_{g}$ as hyperparameters with fixed values, we use the Bayesian estimation to find $\mathcal{U}_{i,j}, \mathcal{V}_{i,j}$ and $\mathcal{S}_{i,j}.$ Based on Bayes rule, we have the following MAP formulation:
\begin{equation}
    \label{Bayesmodel}
    \begin{aligned}
    &p\left(\mathcal{U}_{i,j}, \mathcal{V}_{i,j}, \mathcal{S}_{i,j} | \mathcal{O}_{i,j}, \lambda_{u}, \lambda_{v}, \lambda_{s}, \lambda_{g}\right) \propto \\
    &p\left(\mathcal{O}_{i,j} | \mathcal{U}_{i,j}, \mathcal{V}_{i,j}, \mathcal{S}_{i,j}, \lambda_{g}\right) \\
    &\cdot p\left(\mathcal{U}_{i,j} | \lambda_{u}\right) p\left(\mathcal{V}_{i,j} | \lambda_{v}\right) p\left(\mathcal{S}_{i,j} | \lambda_{s}\right).
    \end{aligned}
\end{equation}
Substituting the distribution of each variable into formula (\ref{Bayesmodel}), and then using the Lemma 6 in \cite{Fnorm2nuclearnorm}, we can get
\begin{equation}
\label{equ:rank_RPCA}
\begin{aligned}
& \arg\min_{\mathcal{L_{i,j}}, \mathcal{S}_{i,j} \in \mathbb{R}^{m_{1}n_{1} \times p}}\|\mathcal{L}_{i,j}\|_{*}+\lambda\|\mathcal{S}_{i,j}\|_{1} \\ & \operatorname{st}\|\mathcal{O}_{i,j}-\mathcal{L}_{i,j}-\mathcal{S}_{i,j}\|_{\text{F}}^{2} \leq \varepsilon, \operatorname{rank}(\mathcal{L}_{i,j}) \leq r.
\end{aligned}
\end{equation}

Because the non-convex $L_{\gamma}$-norm can provide a closer approximation to the matrix rank than the traditional nuclear norm $\|.\|_{*}$,
we use $L_{\gamma}$-norm to represent the LR of hyperspectral data in (\ref{equ:rank_RPCA}) and propose the non-convex local patch-based low-rank model (NonLLR):
\begin{equation}
\label{eq:LTRPCA}
\begin{split}
\displaystyle &\arg\min_{\mathcal{L}_{i,j}, \mathcal{S}_{i,j} }~  \sum_{i,j} \left\|\mathcal{L}_{i,j}\right\|_{\gamma}+\lambda\left\|\mathcal{S}_{i, j}\right\|_{1} \\
&s.t.~\left\|\mathcal{O}_{i, j}-\mathcal{L}_{i, j}-\mathcal{S}_{i, j}\right\|_{\mathrm{F}}^{2} \leq \varepsilon, \operatorname{rank}(\mathcal{L}_{i,j}) \leq r,
\end{split}
\end{equation}
where $\left\|\mathcal{L}_{i, j}\right\|_{\gamma}=\sum_{t=1}^{\min \{m_{1}n_{1}, p\}}\left(1-e^{-\sigma_{t}(\mathcal{L}_{i,j}) / \gamma}\right)$,
$\sigma_{t}(\mathcal{L}_{i,j})$ is the $t$-th singular value of $\mathcal{L}_{i,j}$.

NonLLR (\ref{eq:LTRPCA}) is a local model which exploits the local LR property of HSIs,
while SSTV is a global model which studies the correlations of spatial pixels and spectral bands.
By combining the local low-rank and TV properties in both spatial and spectral domains,
we propose the following NonLLRTV model
\begin{equation}
\label{eq:TLRL1-2LSSTV}
    \begin{split}
  \displaystyle\arg\min_{\mathcal{L},\mathcal{S}} &~ \sum_{i, j}\left(\left\|\mathcal{L}_{i, j}\right\|_{\gamma}+\lambda\left\|\mathcal{S}_{i, j}\right\|_{1}\right)+\tau\|\mathcal{L}\|_ \mathrm{SSTV}  \\
  s.t.&~\left\|\mathcal{O}_{i, j}-\mathcal{L}_{i, j}-\mathcal{S}_{i, j}\right\|_{\mathrm{F}}^{2} \leq \varepsilon, \operatorname{rank}(\mathcal{L}_{i,j}) \leq r,
    \end{split}
\end{equation}
where the SSTV is defined as
\begin{equation}
    \begin{aligned}
\|\mathcal{L}\|_{\text {SSTV }}:=& \sum_{i, j, k} w_{1}\left|l_{i, j, k}-l_{i, j, k-1}\right|+w_{2}\left|l_{i, j, k}-l_{i, j-1, k}\right| \\ &+w_{3}\left|l_{i, j, k}-l_{i-1, j, k}\right|,
    \end{aligned}
\end{equation}
and $w_{1}, w_{2}$ and $w_{3}$ are weighting parameters.

\section{PROPOSED ALGORITHMS}
%The augmented Lagrange multiplier (ALM) method is an  efficient algorithm for solving objective functions with multiple regular terms.
In this section, we use the alternating direction method of multipliers (ADMM) to solve the NonLLRTV model (\ref{eq:TLRL1-2LSSTV}).
Auxiliary variables $\mathcal{J}, \mathcal{X} \in \mathbb{R}^{m \times n \times p}$ are first introduced,
and NonLLRTV model (\ref{eq:TLRL1-2LSSTV}) can be rewritten as follows:
\begin{equation}
\label{equ:model2}
  \begin{split}
    \arg\min_{\mathcal{L}, \mathcal{S}, \mathcal{J}, \mathcal{X}} & \sum_{i, j}\left( \left\|\mathcal{L}_{i, j}\right\|_{\gamma}+\lambda\left\|\mathcal{S}_{i, j}\right\|_{1}\right)+\tau\|\mathcal{X}\|_ \mathrm{SSTV}\\
    s.t.&~\mathcal{J}=\mathcal{X}, \mathcal{L}_{i, j}=\mathcal{J}_{i, j}, \mathcal{U}=\textbf{D}\mathcal{X}\\
    &~\left\|\mathcal{O}_{i, j}-\mathcal{L}_{i, j}-\mathcal{S}_{i, j}\right\|_{\mathrm{F}}^{2} \leq \varepsilon, \operatorname{rank}(\mathcal{L}_{i,j}) \leq r,
  \end{split}
\end{equation}
where $\mathbf{D}=[w_{1}\mathbf{D}_{1}, w_{2}\mathbf{D}_{2}, w_{3}\mathbf{D}_{3}]$ is the forward finite-difference operator along the three modes.
By using ALM and ADMM method, minimization (\ref{equ:model2}) can be transformed into the following two subproblems in Section \ref{sec3.1} and \ref{sec3.2},
where $\Lambda_{i,j}^{\mathcal{O}}$, $\Lambda_{i,j}^{\mathcal{L}}$, $\Lambda_{\mathcal{X}}$ and $\Lambda$ are Lagrangian parameters and $\mu$ is the penalty parameter.

\subsection{Local NonLLR optimization for $(\mathcal{L}, \mathcal{S})$}
\label{sec3.1}
For the above optimization problem, we solve each patch separately and accumulate a weighted sum of $\left(\mathcal{L}_{i, j}, \mathcal{S}_{i, j}\right)$ to reconstruct $(\mathcal{L}, \mathcal{S})$.

Let $\mathcal{T}_{i, j}=\frac{1}{2}\left(\mathcal{O}_{i, j}+\mathcal{J}_{i, j}-\mathcal{S}_{i, j}+\left(\Lambda_{i, j}^{\mathcal{O}}+\Lambda_{i, j}^{\mathcal{L}}\right) / \mu\right)$.
Then, the optimum solution of $\mathcal{L}_{i,j}$-subproblem can be efficiently obtained by the generalized weight singular value thresholding (WSVT) \cite{WSVT}:
\begin{equation}
\label{eq:Lmodel-solution}
\quad \mathbf{L}_{i,j}^{*}=\mathbf{PS}_{\frac{\nabla \phi}{\mu}}(\Sigma) \mathbf{Q}^{\text{T}}
\end{equation}
where $\mathbf{T}_{i,j}=\mathbf{P} \Sigma \mathbf{Q}^{\text{T}}$ is the SVD of $\mathbf{T}_{i,j}$; $\mathbf{S}_{\left(\nabla \phi / \mu\right)}(\Sigma)=$
$\operatorname{diag}\left\{\max \left(\Sigma_{n n}-\left(\nabla \phi\left(\sigma_{n}\right) / \mu\right), 0\right)\right\}$, and $\phi(x)=1-e^{x / \gamma}$.

With $\mathcal{L}_{i, j}$ fixed, the solution of $\mathcal{S}_{i, j}$ can be directly obtained by the soft-thresholding Soft$(\mathcal{M}_{i,j}, \lambda / \mu)$ operation:
\begin{equation}
\label{eq:Smodel-solution}
\mathcal{S}_{i, j}=\operatorname{sign}(\mathcal{M}_{i,j})  \max \left\{0,|\mathcal{M}_{i,j}|-\lambda / \mu\right\},
\end{equation}
where $\mathcal{M}_{i,j}=\mathcal{O}_{i, j}-\mathcal{L}_{i, j}+\Lambda_{i, j}^{\mathcal{O}} / \mu$.

\subsection{Global SSTV regularized reconstruction problem for $(\mathcal{J}, \mathcal{X}, \mathcal{U})$}

The subproblem for $\mathcal{J}$ can be deduced as a convex function, which has the following closed-form solution:
\begin{equation}
\label{eq:Jmodel-solution}
\begin{split}
   \mathcal{J}= & (\mathcal{X}-\Lambda_{\mathcal{X}}/\mu + \sum_{i, j} \operatorname{P}_{i, j}^{\text{T}}\left(\mathcal{L}_{i, j}+\Lambda_{i, j}^{\mathcal{L}} / \mu\right))./ \mathcal{R},
\end{split}
\end{equation}
where $\mathcal{R}=\mathbf{1}+\sum_{i, j} \operatorname{P}_{i, j}^{\text{T}} \operatorname{P}_{i, j}$.

With $\mathcal{J}$ fixed, the subproblem of $\mathcal{X}$ can be efficiently solved by the fast Fourier transform (FFT) method:
\begin{equation}
\label{eq:Xmodel-solution}
\mathcal{X}=\mathcal{F}^{-1}\left[\frac{\mathcal{F}\left(\left(\mathcal{J}+\Lambda_{\mathcal{X}} / \mu\right)+\mathbf{D}^{\text{T}}(\mathcal{U}+\Lambda / \mu)\right)}{1+\sum_{i=1}^{3}\left(\mathcal{F}\left(w_{i} \mathbf{D}_{i}\right)\right)^{2}}\right],
\end{equation}
where $\mathcal{F}(\cdot)$ denotes the FFT and $\mathcal{F}^{-1}$ is the inverse transform;
$\mathbf{D}^{\text{T}}$ represents the adjoint operator of $\mathbf{D}$.

Let $\mathcal{Y}=\left[\mathcal{Y}_{1}, \mathcal{Y}_{2}, \mathcal{Y}_{3}\right]$ and $\mathcal{U}=\left[\mathcal{U}_{1}, \mathcal{U}_{2}, \mathcal{U}_{3}\right]$.
Likewise, $\mathcal{U}$ can be solved by the soft-thresholding operation in (\ref{eq:Smodel-solution}):
\begin{equation}
\mathcal{U}_{t}=\operatorname{Soft}\left(w_{t} \mathbf{D}_{i} \mathcal{X}-\mathcal{Y}_{t} / \mu, \tau / \mu\right), t=1,2,3.
\label{eq:Umodel-solution}	
\end{equation}

\subsection{Updating Lagrangian parameters $\Lambda_{i,j}^{\mathcal{O}}$, $\Lambda_{i,j}^{\mathcal{L}}$ and $\Lambda_{\mathcal{X}}$}
\label{sec3.2}
\begin{equation}
\label{eq:Lagparas}
\left\{
\begin{split}
 & \Lambda_{i, j}^{\mathcal{O}}=\Lambda_{i, j}^{\mathcal{O}}+\mu\left(\mathcal{O}_{i, j}-\mathcal{L}_{i, j}-S_{i, j}\right),\\
 & \Lambda_{i, j}^{\mathcal{L}}=\Lambda_{i, j}^{\mathcal{L}}+\mu\left(\mathcal{L}_{i, j}-\mathcal{J}_{i, j}\right), \\
 & \Lambda_{\mathcal{X}}=\Lambda_{\mathcal{X}}+\mu(\mathcal{J}-\mathcal{X}),\\
 & \Lambda=\Lambda+\mu(\mathcal{U}-\mathbf{D}\mathcal{X}).
\end{split}
\right.
\end{equation}

\renewcommand{\algorithmicrequire}{\textbf{Input:}} % Use Input in the format of Algorithm
\renewcommand{\algorithmicensure}{\textbf{Output:}} % Use Output in the format of Algorithm

Algorithm \ref{algorightm-1} summarizes the optimization strategy of step-by-step iteration as above.
\begin{algorithm}
	\caption{HSI restoration via NonLLRTV model.} \label{algorightm-1}
\begin{algorithmic}[1]
    \Require
      $m \times n \times p$ observed HSI $ \mathcal{O}$, patch size $m_1 \times n_1$, stopping criterion $\varepsilon$, regularization parameters $\lambda$, $\tau$, $\gamma$.
    \Ensure
      Denoised image $\mathcal{X}$;
    \State Initialize: $\mathcal{L}=\mathcal{X}=\mathcal{S}=\mathcal{J}=0,$
    $\Lambda_{i, j}^{\mathcal{O}}=\Lambda_{i, j}^{\mathcal{L}}=0$, $\Lambda_{\mathcal{X}}=0$, $\Lambda=0$, $\mu=10^{-2}, \mu_{\max }=10^{6}, \rho=1.5, \gamma \in (7 * 10^{-3}, 1.2 * 10^{-2})$
   ; $w_{1}=w_{2}=1, w_{3}=0.5;$ $\lambda=0.14, \tau=0.03$ and $k=0$.
    \State Update all patches $\left(\mathcal{L}_{i, j}, \mathcal{S}_{i, j}\right)$  by (\ref{eq:Lmodel-solution}) and (\ref{eq:Smodel-solution})
    respectively;
    \State Update $\mathcal{J}$, $\mathcal{X}$ , $\mathcal{U}$ by (\ref{eq:Jmodel-solution}), (\ref{eq:Xmodel-solution}), (\ref{eq:Umodel-solution}) respectively;
    \State Update the Lagrangian multipliers by (\ref{eq:Lagparas});
    \State Update the penalty parameter by $\mu :=\min \left(\rho \mu, \mu_{\max }\right)$
    \State Check the convergence condition:\\
    $\max \left\{\left\|\mathcal{O}_{i, j}-\mathcal{L}_{i, j}^{k+1}-\mathcal{S}_{i, j}^{k+1}\right\|_{\infty},\left\|\mathcal{J}^{k+1}-\mathcal{X}^{k+1}\right\|_{\infty}\right\}\leq \varepsilon.$
\end{algorithmic}
\end{algorithm}

\begin{table*}
    \caption{Quantitative evaluation of different methods in different noise cases of Indian Pines dataset}
	\centering
    \label{tab:PQIs}
\begin{tabular}{p{1cm}lllllll}
	\hline \hline
	Noise& Evaluation index & BM3D \cite{BM3D} & NAILRMA \cite{NAILRMA} & LRMR \cite{LRMR} & LRTV \cite{LRTV} &  NonLLRTV  \\
	\hline \hline
	\multirow{1}{*}{Case 1} & MPSNR/ MSSIM & 28.676/ \underline{0.945}  & 24.295/ 0.768  & 33.757/ 0.892  & \underline{34.497}/ 0.886  & \textbf{37.564/ 0.982}\\
	\hline
	\multirow{1}{*}{Case 2} & MPSNR/ MSSIM & 28.779/ \underline{0.946} & 28.190/ 0.841 & 33.986/ 0.893 & \underline{35.642}/ 0.904 & \textbf{38.805/ 0.986}\\
	\hline
	\multirow{1}{*}{Case 3} & MPSNR/ MSSIM & 29.277/ \underline{0.949}  & \underline{37.144}/ 0.937 & 35.320/ 0.911 & 35.854/ 0.903 & \textbf{40.219/ 0.989}\\
	\hline
	\multirow{1}{*}{Case 4} & MPSNR/ MSSIM & 28.718/ \underline{0.946} & 28.089/ 0.841 & 33.679/ 0.891 & \underline{35.258}/ 0.899 & \textbf{38.435/ 0.985}\\
	\hline
	\multirow{1}{*}{Case 5} & MPSNR/ MSSIM & 28.647/ \underline{0.946} & 27.467/ 0.826 & 33.523/ 0.890 & \underline{34.854}/ 0.910 & \textbf{38.394/ 0.986}\\
	\hline
	\multirow{1}{*}{Case 6} & MPSNR/ MSSIM & 28.573/ \underline{0.945} & 27.412/ 0.831 & 33.207/ 0.886 & \underline{34.271}/ 0.900 & \textbf{38.131/ 0.985}\\
	\hline \hline
\end{tabular}
\end{table*}

\section{EXPERIMENTS AND RESULTS}
\label{results}

In order to verify the effectiveness of our proposed model for HSI restoration, four different methods are employed as the benchmark in the experiments.
Since the BM3D is only suitable to remove Gaussian noise, we implement them on HSIs which are preprocessed by the classical RPCA restoration method. The classical Indian Pines dataset is selected to apply simulated experiments. The parameters of the proposed model are given in Algorithm 1. To simulate noisy HSI data, we add several types of noise to the original HSI data.

Case 1: Gaussian noise and impulse noise are added to the HSI. The mean value of Gaussian noise is zero and the variance is 0.05, the percentage of impulse noise is 0.1;

Case 2: Noise type is the same as Case 1, the mean value of Gaussian noise is zero, while its variance and the percentage of impulse noise for each band is randomly selected from 0 to 0.2;

Case 3: Only Gaussian noise is added to the HSI, the mean value of Gaussian noise is zero and the variance for each band is randomly selected from 0 to 0.2;

Case 4: Based on Case 2, deadlines are additionally added from band 131 to band 160;

Case 5: Based on Case 2, stripes are additionally added from band 111 to band 140;

Case 6: Based on Case 2, the deadlines and stripes in Case 4 and Case 5 both are added to the HSI.

\begin{figure*}[!t]
	\centering
%	\includegraphics[width=0.35\linewidth]{myfigure/SSIM/tuli2}\\
%	\label{*}
	\subfloat[True image]{\includegraphics[width=0.13\linewidth]{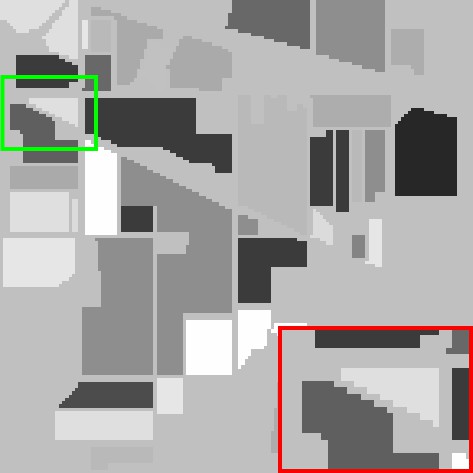}}%
	\hfil
	\subfloat[Noisy (a)]{\includegraphics[width=0.13\linewidth]{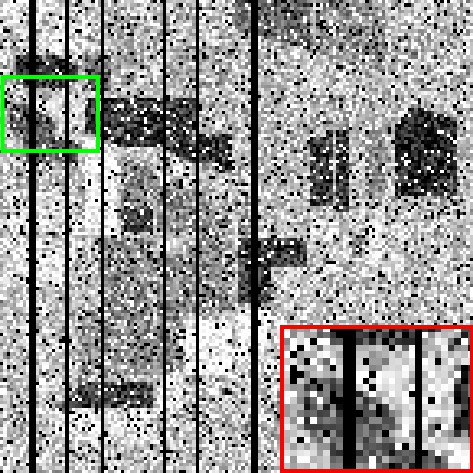}}%
	\label{*}
	\hfil
	\subfloat[BM3D]{\includegraphics[width=0.13\linewidth]{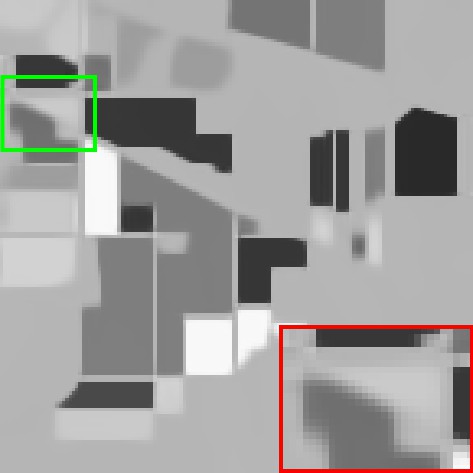}}%
	\label{*}
	\hfil
	\subfloat[NAILRMA]{\includegraphics[width=0.13\linewidth]{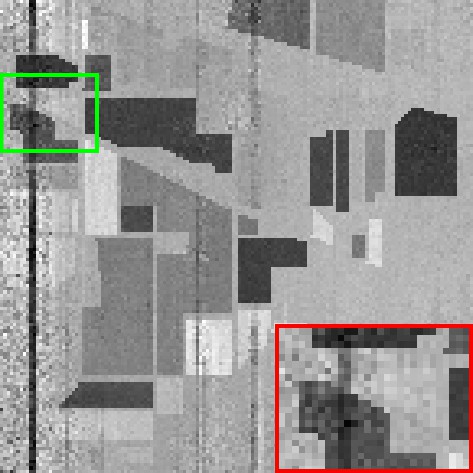}}%
	\label{*}
	\hfil
	\subfloat[LRMR]{\includegraphics[width=0.13\linewidth]{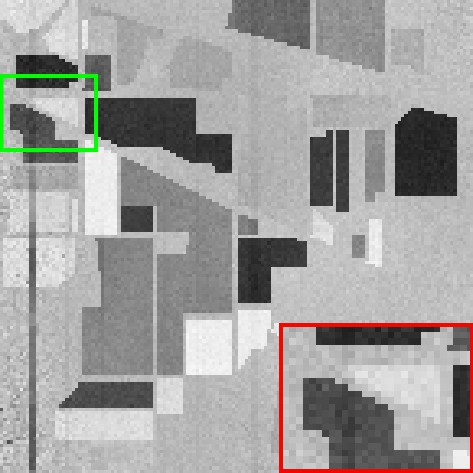}}%
	\label{*}
	\hfil
	\subfloat[LRTV]{\includegraphics[width=0.13\linewidth]{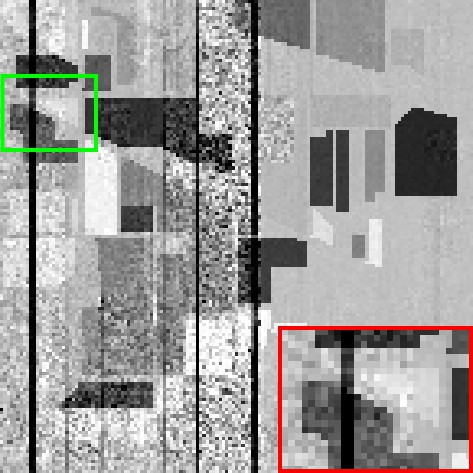}}%
	\label{*}
	\hfil
	\subfloat[NonLLRTV]{\includegraphics[width=0.13\linewidth]{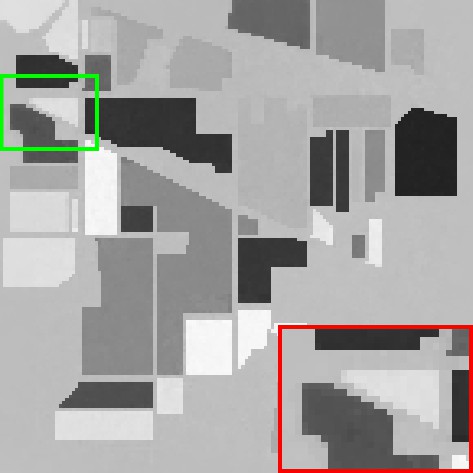}}%
	\label{*}
	\caption{Restoration results in the $140$-th band of Indian Pines dataset in Case 4.
     The PSNRs of (c)-(g) are 25.9026, 19.3116, 25.9038, 16.1044 and 35.1247, respectively.  }
	\label{fig:pgd_band140_indian}
\end{figure*}

\begin{figure*}[!t]
	\centering
%	\includegraphics[width=0.35\linewidth]{myfigure/SSIM/tuli2}\\
%	\label{*}
	\subfloat[True image]{\includegraphics[width=0.132\linewidth]{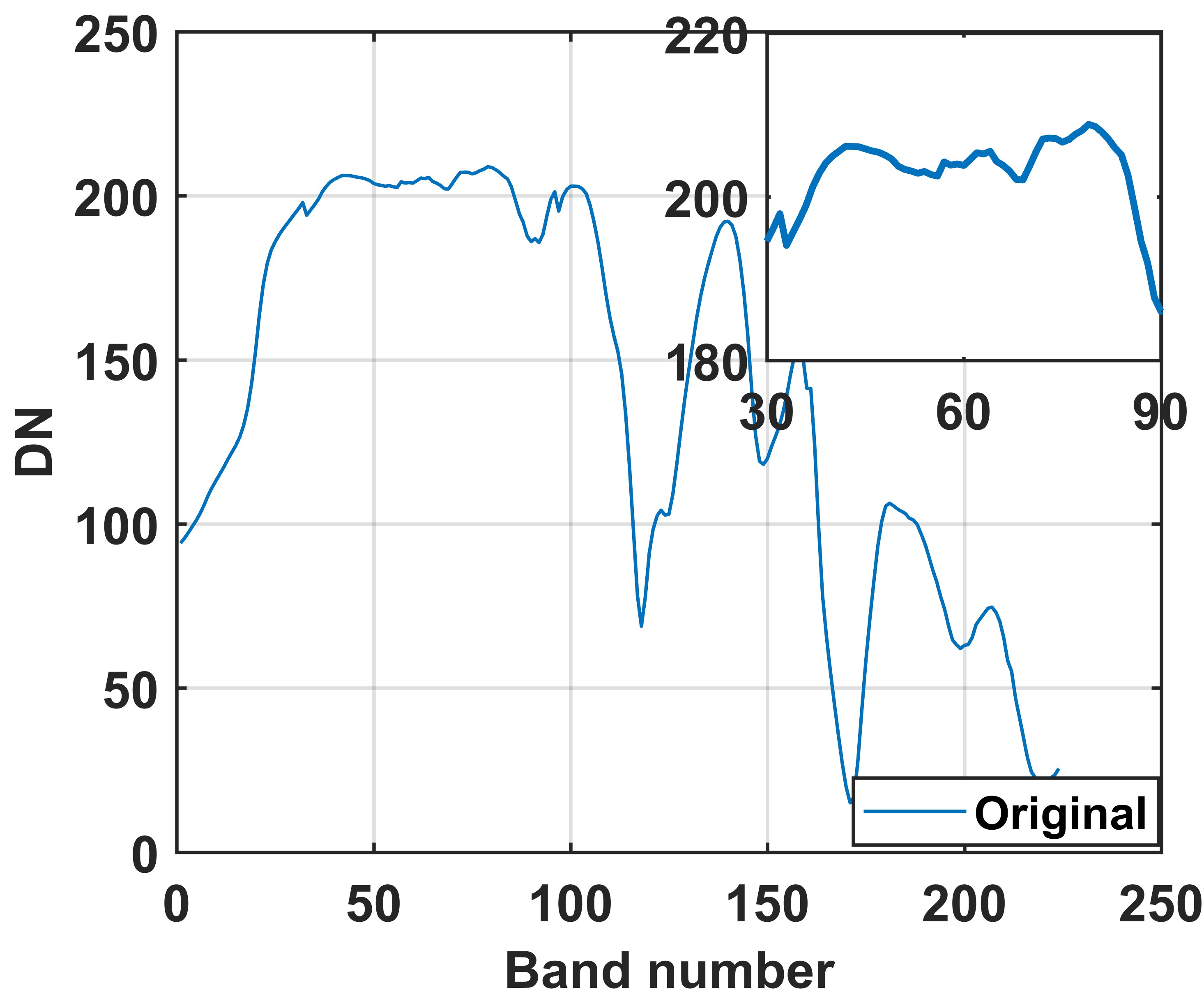}}%
	\hfil
	\subfloat[Noisy (a)]{\includegraphics[width=0.132\linewidth]{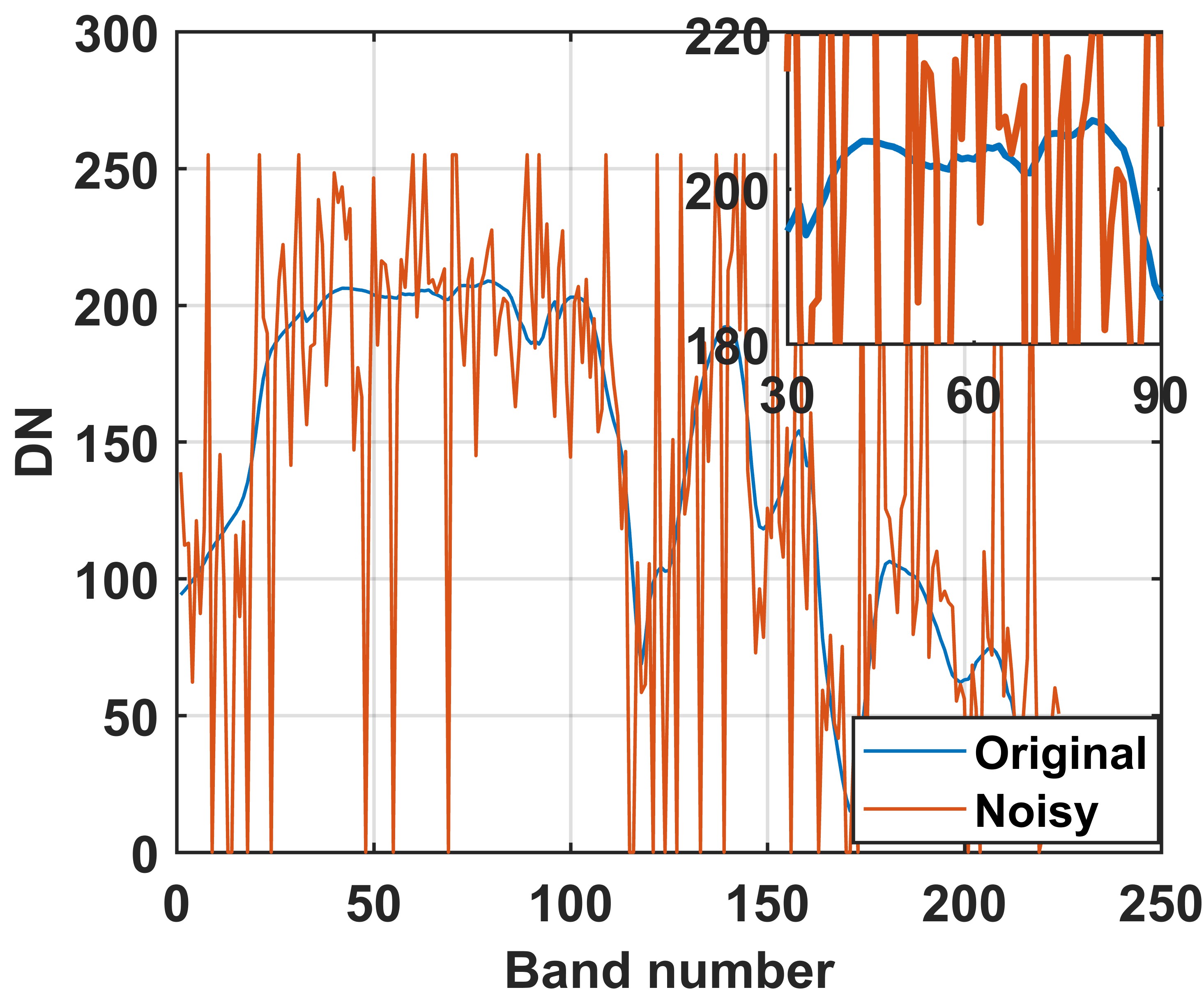}}%
	\label{*}
	\hfil
	\subfloat[BM3D]{\includegraphics[width=0.132\linewidth]{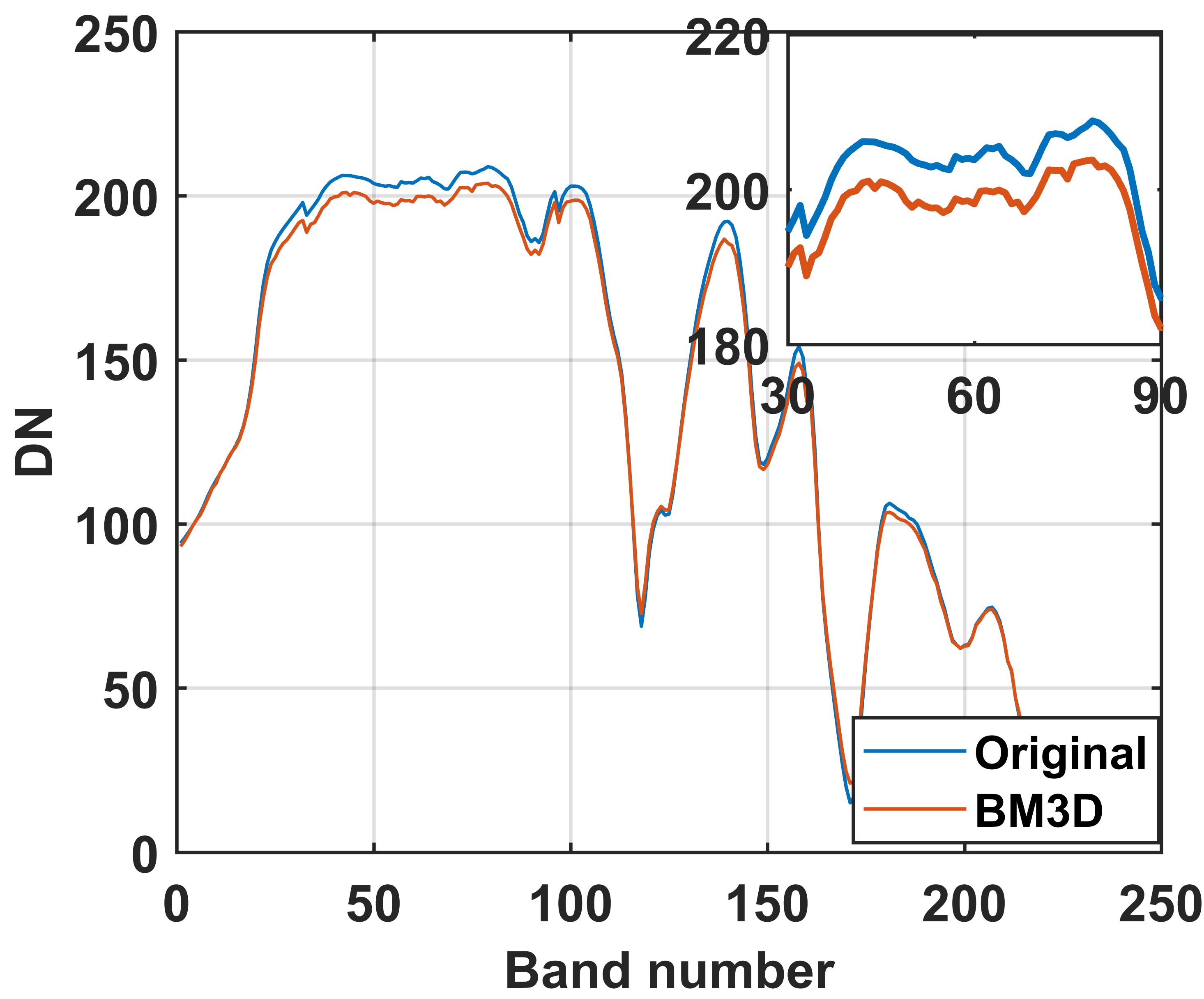}}%
	\label{*}
	\hfil
	\subfloat[NAILRMA]{\includegraphics[width=0.132\linewidth]{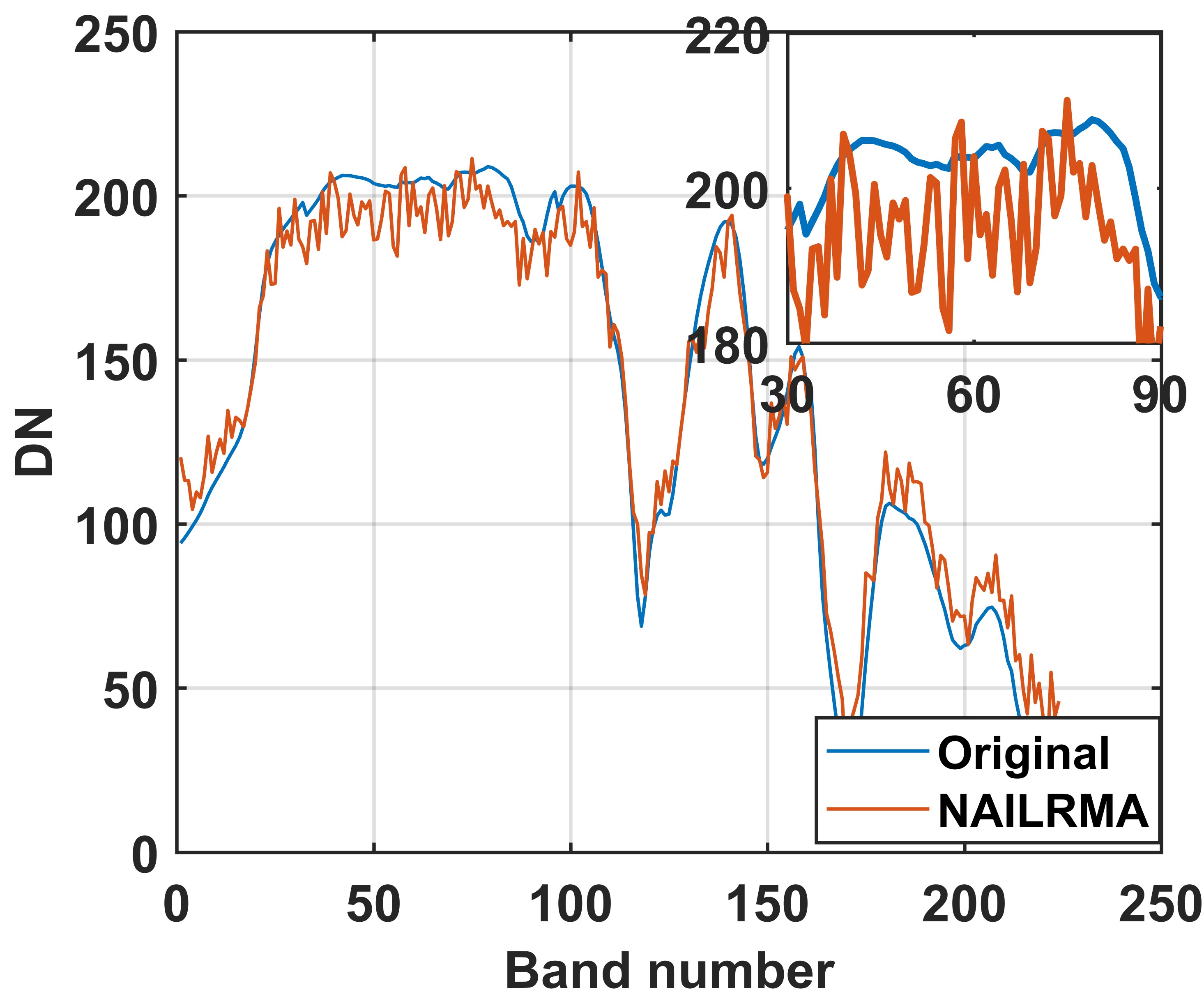}}%
	\label{*}
	\hfil
	\subfloat[LRMR]{\includegraphics[width=0.132\linewidth]{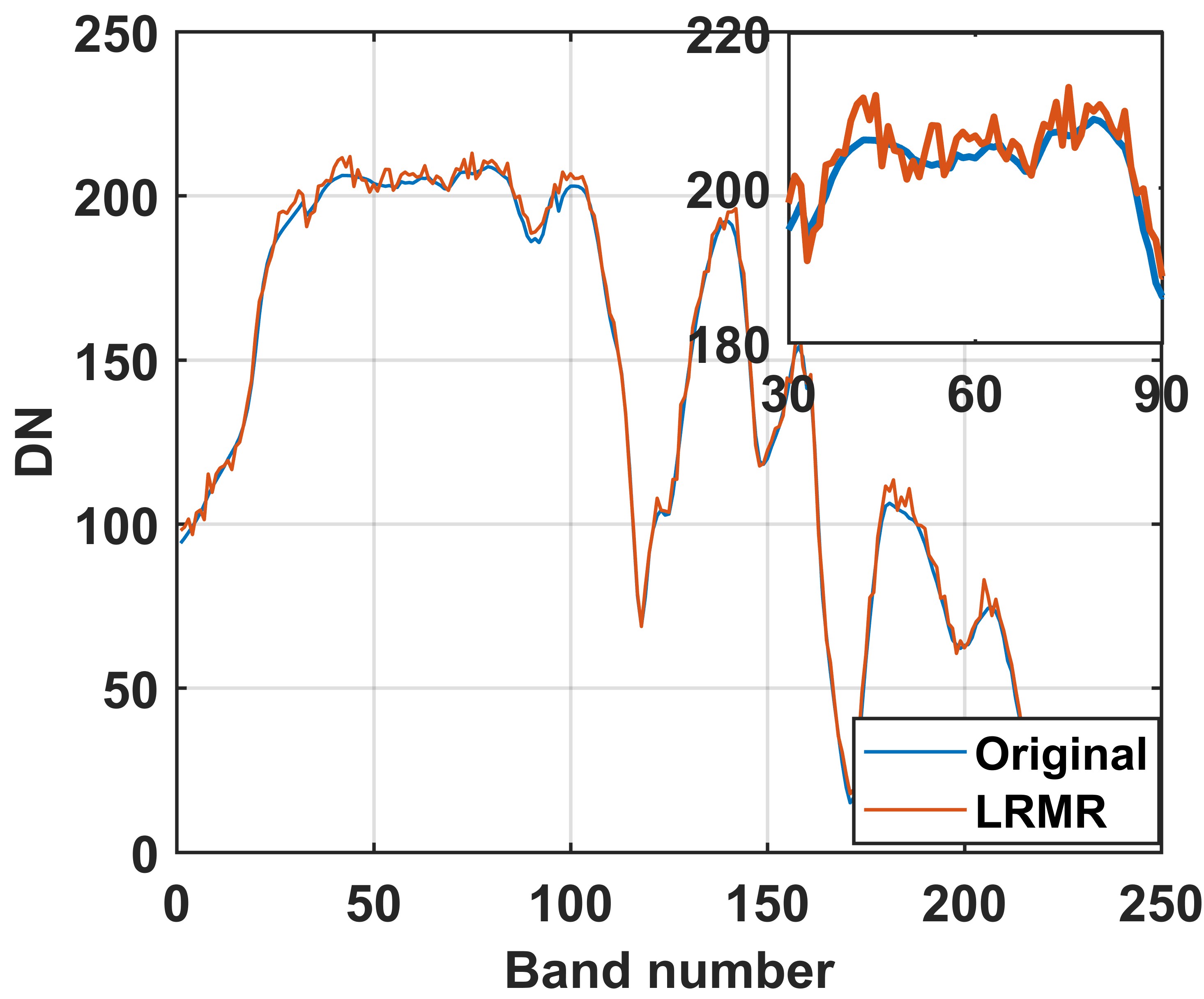}}%
	\label{*}
	\hfil
	\subfloat[LRTV]{\includegraphics[width=0.132\linewidth]{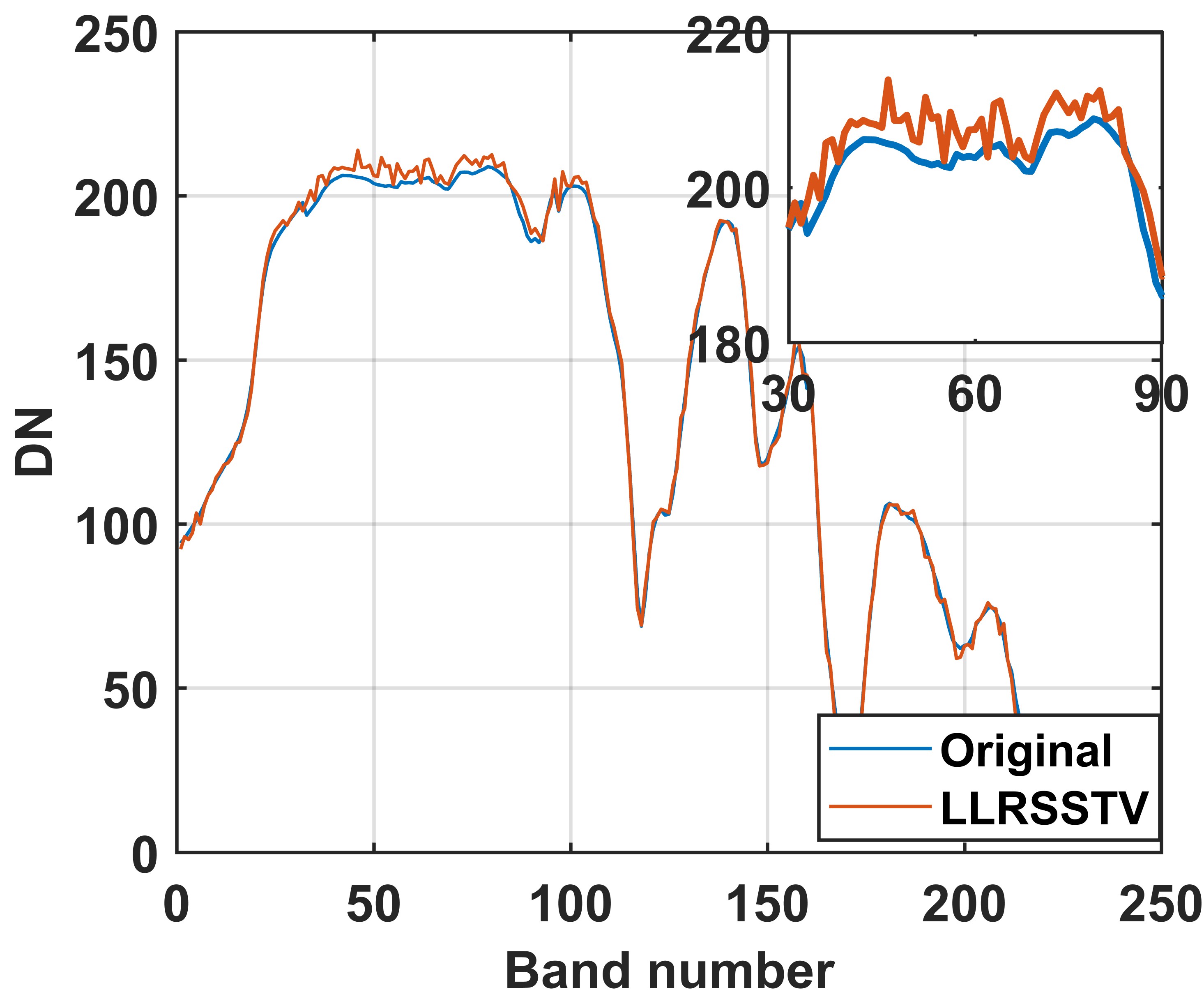}}%
	\label{*}
	\hfil
	\subfloat[NonLLRTV]{\includegraphics[width=0.132\linewidth]{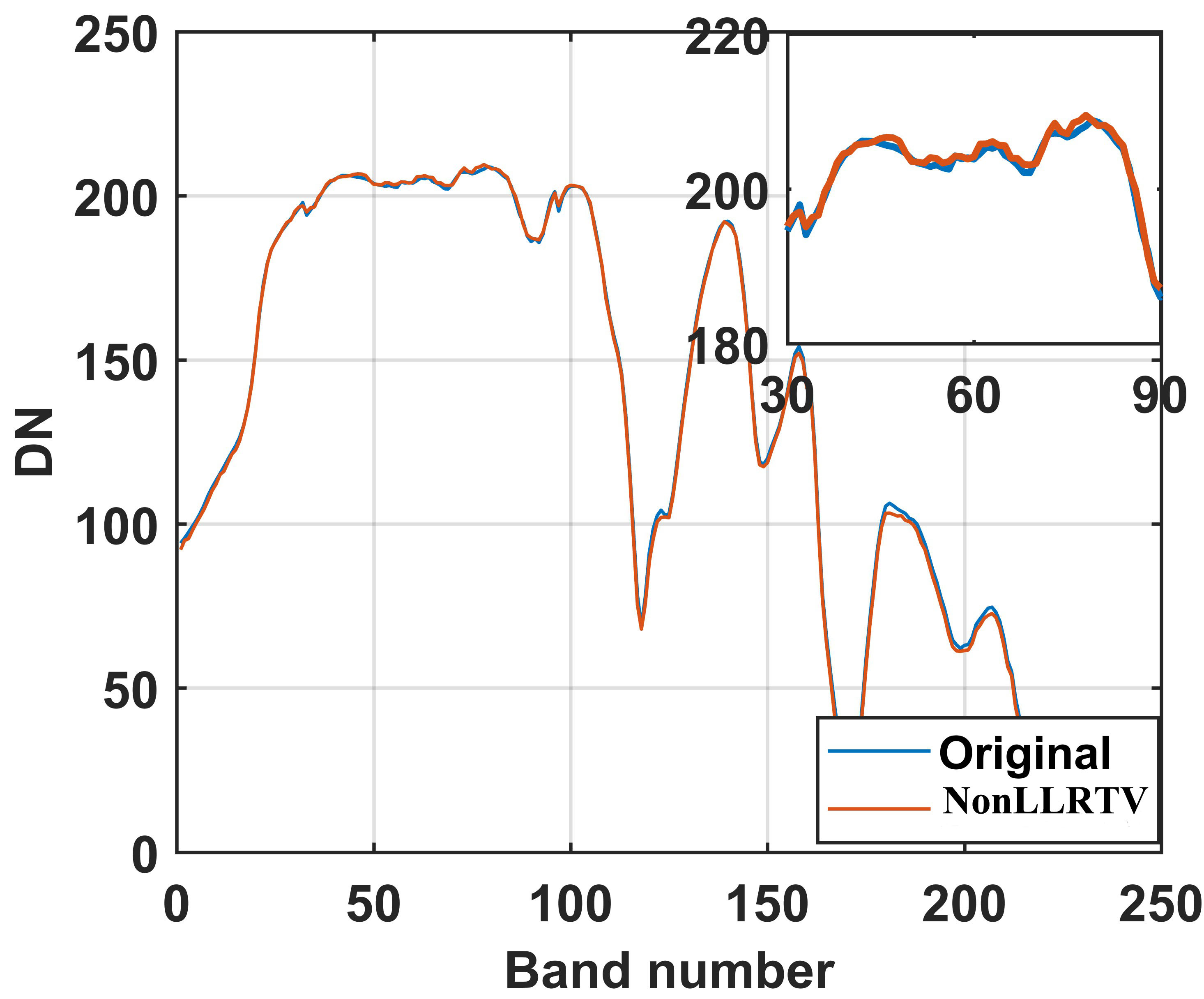}}%
	\label{*}
	\caption{Spectrum of pixel (110, 110) in the restoration results of Indian Pines dataset in Case 1.}
%     (a) True image, (b) Noisy image, (c) BM3D,
%    (d) NAILRMA, (e) LRMR, (f) LRTV, (g) NonLLRTV.}
	\label{fig:Spectrum_p02g01_indian}
\end{figure*}

For visual evaluation, we show the 140-th band of the recovered HSI with the noise case 4 in Fig. \ref{fig:pgd_band140_indian}. Compared to other models,
it can be seen that the result of our NonLLRTV model is closest to the original reference image. In addition, in order to further compare the performance of the models, we show the spectral characteristics of the clean HSI and the restored HSI in Fig. \ref{fig:Spectrum_p02g01_indian}. It is also clear that the spectral characteristics in results
of our NonLLRTV model are also closest to ones in the true data.

For quantitative comparison, Table \ref{tab:PQIs} lists the PQIs of all the compared models in the six noise cases.
The best results for each PQI are marked in bold.
It is clear from Table 1 that in all cases our NonLLRTV model achieves the best results among all the test methods.
It is worth noting that the NonLLRTV model is about 3.5 dB better in MPSNR compared to the suboptimal method.

\section{conclusion} \label{conclusion}

In this paper, we investigated the restoration of HSI, and propose a spatial–spectral TV regularized nonconvex local
LR matrix approximation (NonLLRTV) method to remove mixed noise in HSIs.
The mixed noise was simulated by using various combinations of Gaussian noise, impulse noise, stripes, and deadlines. Results on the Indian Pines dataset indicate that
the use of local nonconvex penalty and global SSTV can boost the preserving of spatial piecewise
smoothness and overall structural information. Our future steps include the investigation of
other types and combinations of nonconvex penalties. Moreover, we will also investigate the use of these models on other high-dimensional data recovery.

% Below is an example of how to insert images. Delete the ``\vspace'' line,
% uncomment the preceding line ``\centerline...'' and replace ``imageX.ps''
% with a suitable PostScript file name.
% -------------------------------------------------------------------------

%\vfill
%\pagebreak

\bibliographystyle{IEEEbib}
\bibliography{refs}

\end{document}